\documentclass[10pt,twocolumn,letterpaper]{article}
\usepackage{iccv}              
\usepackage{multirow}
\usepackage{array}
\usepackage{graphicx}
\usepackage{booktabs}
\usepackage{makecell}
\usepackage{amsmath}
\usepackage{caption}
\usepackage{tikz}
\usepackage{amssymb}
\usepackage{bbding}
\usepackage{url}
\usepackage[table]{xcolor} 

\usetikzlibrary{fadings}
\definecolor{iccvblue}{rgb}{0.21,0.49,0.74}
\usepackage[pagebackref,breaklinks,colorlinks,allcolors=iccvblue]{hyperref}

\def\paperID{12521} 
\def\confName{ICCV}
\def\confYear{2025}

\author{
Tao Han\textsuperscript{1,2}, \quad Wanghan Xu\textsuperscript{2}, \quad Junchao Gong\textsuperscript{2}, \quad Xiaoyu Yue\textsuperscript{2,3},\\
Song Guo\textsuperscript{1\Envelope}, \quad Luping Zhou\textsuperscript{3}, \quad Lei Bai\textsuperscript{2\Envelope}\\
\textsuperscript{1}Hong Kong University of Science and Technology \quad
\textsuperscript{2}Shanghai Artificial Intelligence Laboratory \\
\textsuperscript{3}The University of Sydney \\
{\tt\small hantao10200@gmail.com, \{xuwanghan, gongjunchao, yuexiaoyu\}@pjlab.org.cn} \\ {\tt\small songguo@ust.hk, luping.zhou@sydney.edu.au, bailei@pjlab.org.cn}
}

\newcommand{\lred}[1]{\textcolor{red!50!white}{(#1)}}

\title{%
    \includegraphics[width=1.5em]{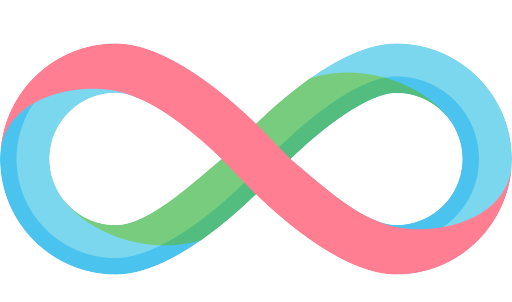}~InfGen: A Resolution-Agnostic Paradigm for Scalable Image Synthesis%
}

\begin{document}
\twocolumn[{%
\renewcommand\twocolumn[1][]{#1}%
\maketitle
\begin{center}
    \centering
    \captionsetup{type=figure}
    \includegraphics[width=0.95\linewidth]{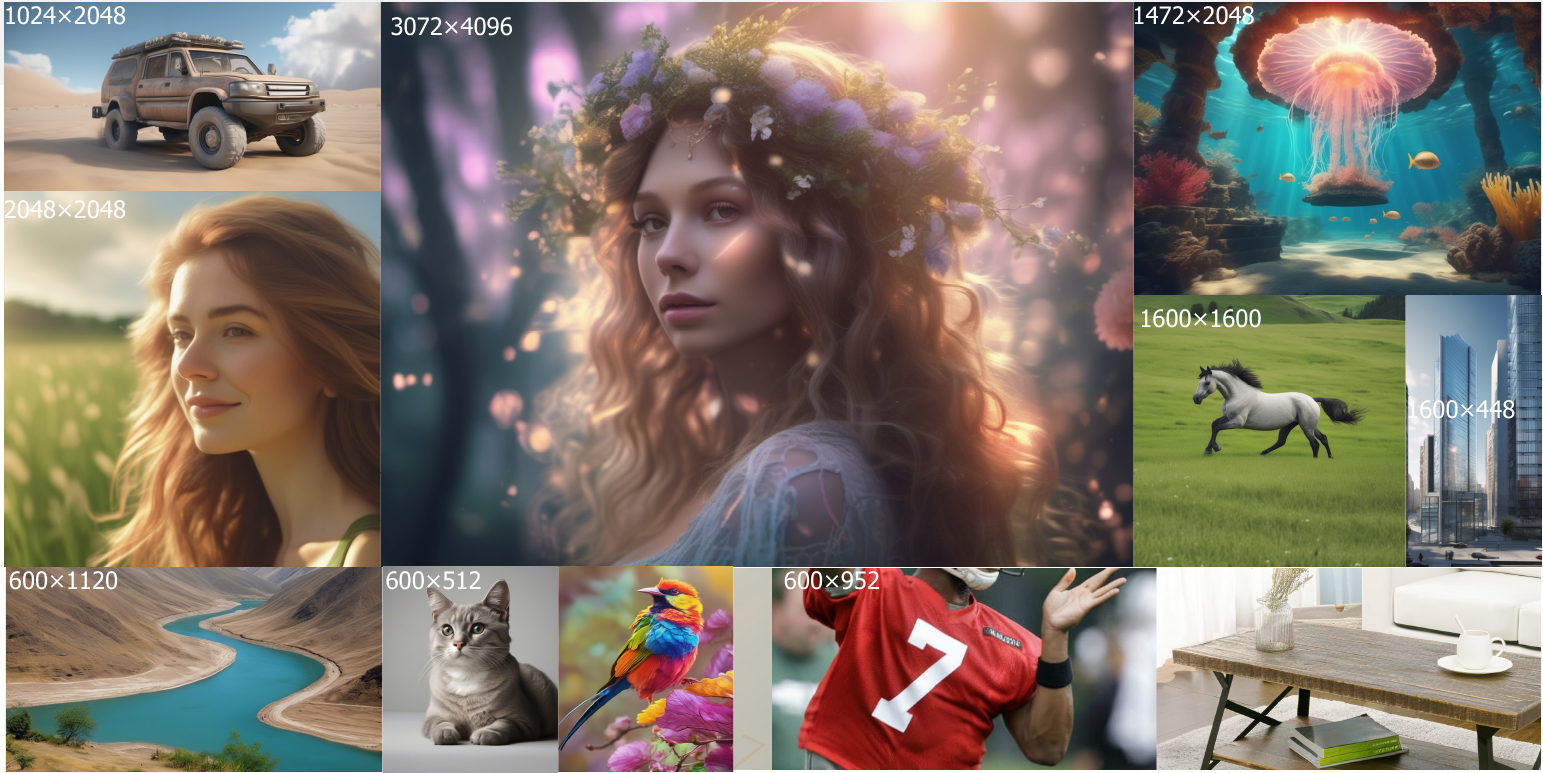}
    \captionof{figure}{The proposed InfGen creates highly photo-realistic and detail-rich images at various resolutions when it is applied on SDXL~\cite{podell2023sdxl}. Best viewed zoomed in. For more image generation, please visit our demo website to experience it.}
    \label{fig:placeholder}
\end{center}%
}]

\renewcommand{\thefootnote}{}
\footnotetext{\Envelope \hspace{2pt} Corresponding Authors.}
\renewcommand{\thefootnote}{\arabic{footnote}} 

\begin{abstract}
Arbitrary resolution image generation provides a consistent visual experience across devices, having extensive applications for producers and consumers. Current diffusion models increase computational demand quadratically with resolution, causing 4K image generation delays over 100 seconds. To solve this, we explore the second generation upon the latent diffusion models, where the fixed latent generated by diffusion models is regarded as the content representation and we propose to decode arbitrary resolution images with a compact generated latent using a one-step generator. Thus, we present the \textbf{InfGen}, replacing the VAE decoder with the new generator, for generating images at any resolution from a fixed-size latent without retraining the diffusion models, which simplifies the process, reducing computational complexity and can be applied to any model using the same latent space. Experiments show InfGen is capable of improving many models into the arbitrary high-resolution era while cutting 4K image generation time to under 10 seconds\footnote{\url{https://github.com/taohan10200/InfGen}}.
\end{abstract}

\vspace{-12pt}

\section{Introduction}
\label{sec:intro}
In recent years, image generation has been significantly advanced by diffusion models~\cite{rombach2022high,peebles2023scalable,chang2022maskgit,podell2023sdxl} and autoregressive models~\cite{sun2024autoregressive,chang2022maskgit}.
Due to the high training costs and slow multi-step inference, most generative methods follow a two-stage paradigm: First, the generative model produces an intermediate representation of the image, and then the decoder from a pre-trained tokenizer maps the intermediate representation to pixels.
The first stage typically employs generative models, such as LDMs~\cite{rombach2022high} and DiT~\cite{peebles2023scalable}, to generate image representations, while the second stage uses autoencoder models such as VAE~\cite{kingma2013auto,rezende2014stochastic} and VQGAN~\cite{yu2021vector} to decode the representations.

\begin{figure}[t]
    \centering
    \includegraphics[width=1.00\linewidth]{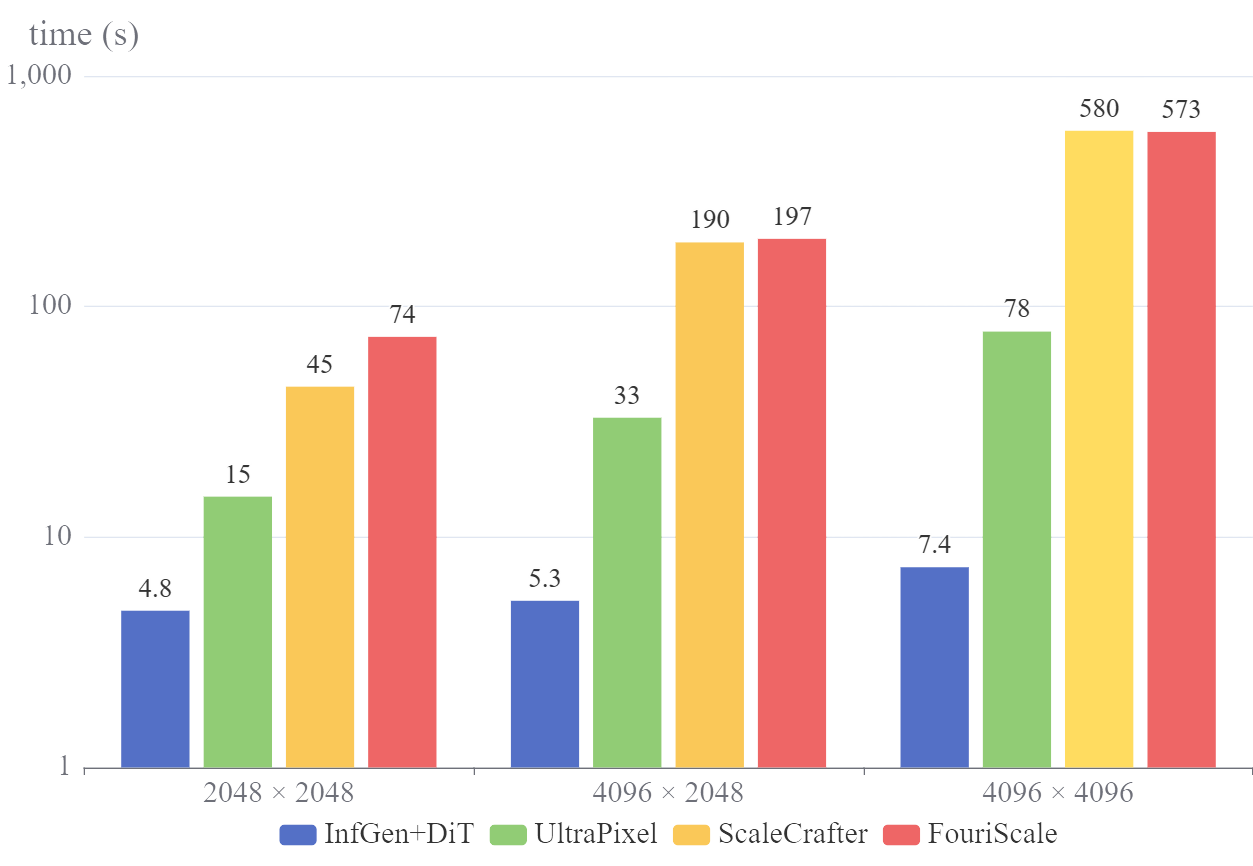}
    \caption{\textbf{Inference time} (seconds per image) for high-resolution image generation methods. The vertical axis is a logarithmic scale.}
    \label{fig:infer_time}
\end{figure}

The research community has long focused on the generative models of the first stage, neglecting the second stage of mapping latents to pixels.
However, modern generative models have their own challenges.
For tasks such as arbitrary high-resolution image generation, diffusion models typically require thousands of sampling steps, leading to prohibitively high computational costs.
For instance, generating a single $4K$ image with a diffusion model can take over $100$ seconds~\cite{huang2024fouriscale, he2023scalecrafter, yang2024inf}. 
In contrast, the decoder in the second stage typically requires only a single forward pass. Therefore, enhancing the capabilities of the decoder to achieve efficient high-resolution image generation is a more effective approach.

In this work, we shift our attention to the mapping from intermediate representations to image pixel values.
A tokenizer is essentially an autoencoder, where the encoder compresses the input image into a more compact latent space to facilitate the design of the generator, while the decoder performs the inverse operation, reconstructing images from generated samples in the latent space.
It is evident that by mapping latent vectors to higher-resolution images during the decoding process, the arbitrary high-resolution image generation task can be simplified into the super-resolution task, which has already been well addressed.
However, reconstructing low-resolution (\eg{} $32 \times 32$ or $64 \times 64$) latent feature maps to high-resolution images (\eg{} $4K$) remains a challenging task.

To this end, we propose generating high-resolution images by conditioning on generated latent maps, resulting in a simple yet effective framework, InfGen.
Since the image content has already been obtained by the diffusion model, we can employ a lightweight generator to capture fine-grained textures and details.
InfGen is a decoder-based generator, which embeds a vision transformer decoder to transfer generated latent into any resolution through cross-attention operation conditioned on the latent. Notably, although VAE employs perceptual loss and GAN loss during training, it still lacks generative capability like InfGen. The reason is that the reconstruction target in the VAE is in the same pixel space as the input, meaning the required information for restoring the target is complete. However, the size of the reconstructed target in InfGen is greater or much bigger than the input image; thus, it is required to have generative ability to complete information loss.

InfGen is trained on a widely used VAE model, 
which is adopted by Stable Diffusion~\cite{rombach2022high,podell2023sdxl}, DiT~\cite{peebles2023scalable}, and SiT~\cite{ma2024sit}.
Therefore, our InfGen acts as a program patch, upgrading these methods to high-resolution image generation with a low computational cost. ~\cref{fig:placeholder} shows images generated by off-the-shelf generative models, which show InfGen can improve the resolution of SD-1.5~\cite{rombach2022high} into arbitrary resolution diffusion model while keeping photo-realistic generation ability. Notably,  our proposed paradigm for generating arbitrary image resolutions offers significant advantages in inference latency, improving the generation speed for images larger than 4K by \textcolor{red}{$10 \times$} compared to the previously fastest UltraPixel~\cite{ren2024ultrapixel} as shown in~\cref{fig:infer_time}.


In summary, our main contributions are:
\begin{itemize}
  \item \textbf{New Paradigm.} Introducing a novel paradigm for generating images at arbitrary resolutions. The secondary generation based on generated latent is an unexplored area.
  \item \textbf{Plug and Play.} The generator can serve as a plugin to upgrade all models based on VAE without further training, showing significant improvements in empowering the existing generative models for arbitrary resolutions.
  \item \textbf{High Quality and Fast.} Compared to existing state-of-the-art methods, the proposed approach not only achieves top-tier generation quality but also improves generation speed by over ten times.
\end{itemize}

\section{Related Work}
\subsection{Latent diffusion image generation.}
Diffusion models~\cite{sohl2015deep,song2020denoising,ho2020denoising,dhariwal2021diffusion,nichol2021glide} are a type of generative model commonly used for producing high-quality images. They function by establishing a Markov diffusion chain, which gradually adds Gaussian noise during the diffusion process and learns to remove this noise using neural networks. In the generation phase, the model samples random Gaussian noise and iteratively generates a clear image through a denoising process. 

The latent diffusion models~\cite{rombach2022high,podell2023sdxl,ramesh2022hierarchical,saharia2022photorealistic,Zheng2024MaskDiT,lu2024fit} utilizes an autoencoder, such as VAE~\cite{kingma2013auto,rezende2014stochastic, xu2025exploring} or VQVAE~\cite{van2017neural,razavi2019generating}, to encode images from pixel space to latent space, conducting diffusion training and inference within this latent space. Finally, the corresponding decoder restores the latents to pixel space, generating the final image. This approach leverages the encoder's compression, enabling efficient training and inference.

\subsection{High-resolution image generation.}

High-resolution image generation has gained popularity in recent years. A common approach is to first generate low-resolution images and then use a super-resolution methods~\cite{zhang2021designing,dong2015image,liang2021swinir,wang2021real} to upscale them. However, super-resolution in pixel space is labor-intensive and underutilizes the decoder's generation capacity. Some methods modify the diffusion inference process to achieve super-resolution. For instance, ScaleCrafter~\cite{he2023scalecrafter} adjusts image resolution during inference by dilating convolutions, while FouriScale~\cite{huang2024fouriscale} employs dilated convolution in the frequency domain to enhance resolution. Additionally, another method~\cite{jin2023training} modifies attention entropy in the denoising network's attention layer based on feature resolutions. However, these training-free methods are often closely tied to specific network architectures and inference processes, limiting their applicability to particular generative models and reducing their generalizability to others.

In addition to training-free methods, some approaches achieve high-resolution generation by redesigning network structures. For example, Inf-DiT~\cite{yang2024inf} employs a unidirectional block attention mechanism to adaptively manage memory overhead during inference. UltraPixel~\cite{ren2024ultrapixel} incorporates Guidance Fusion and Scale-Aware Norm between the blocks of the diffusion model, enabling high-resolution image generation through fine-tuning.

\section{Method}
\paragraph{Overview.} Diffusion models using the $\epsilon$-prediction paradigm typically operate in latent space. Generating arbitrary-resolution images involves handling latents of varying shapes, resulting in high computational costs and latency for ultra-high-resolution generation. This paper proposes a faster approach for arbitrary-resolution generation with diffusion models. In Section \ref{Sec:preliminary}, we redefine arbitrary resolution generation as a two-stage task. Without modifying the diffusion model, we replace the VAE decoder with a generator capable of decoding at arbitrary resolutions. Section \ref{sec: Arbitrary Image Out} details the design and training of such a generator. Since training high-resolution decoders is computationally expensive, Section \ref{sec: training} introduces an iterative extrapolation scheme for ultra-high-resolution generation, offering a training-free resolution enhancement method.

\begin{figure*}
    \centering
    \includegraphics[width=0.99\linewidth]{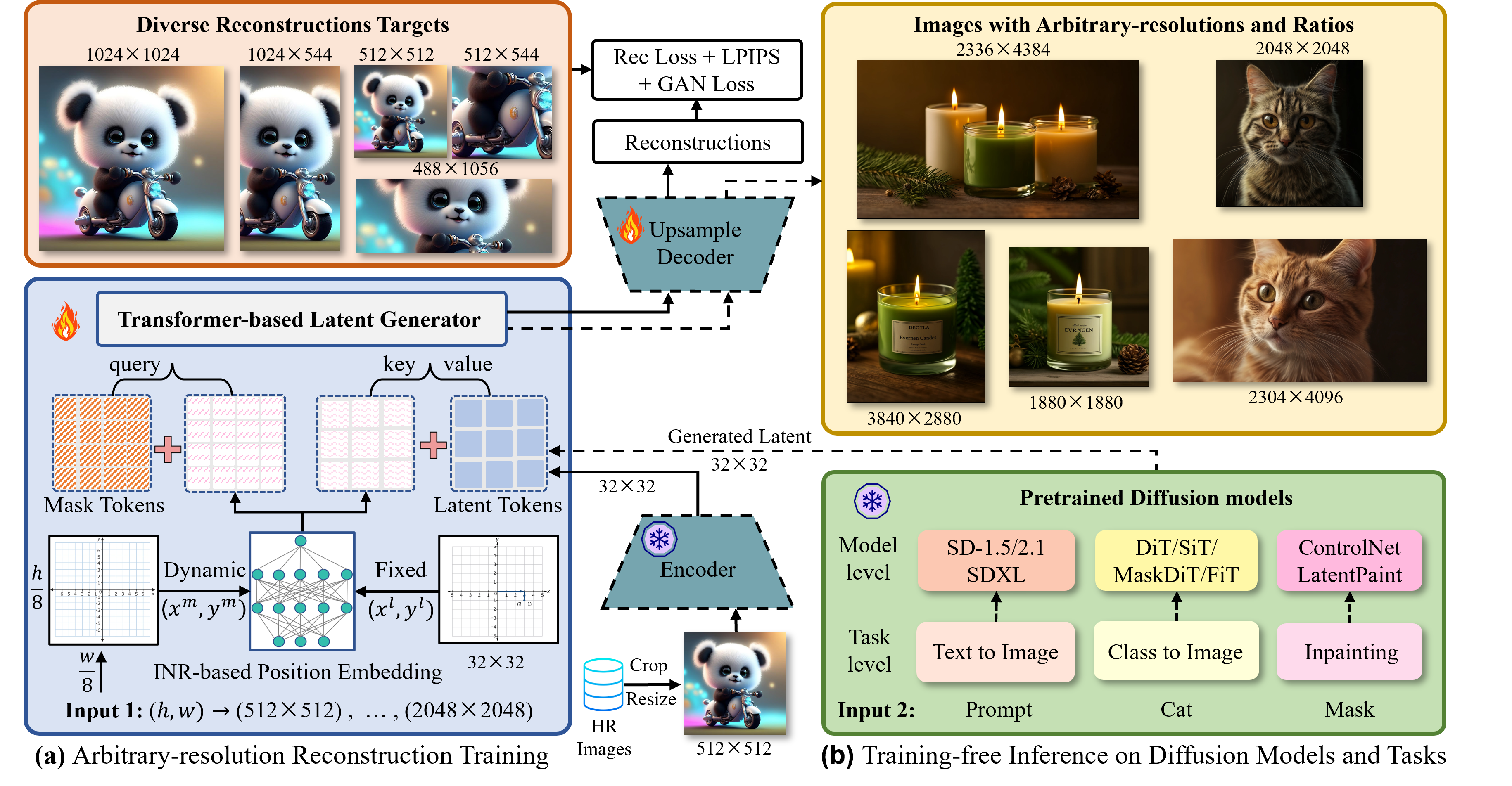}
    \caption{\textbf{Illustration of the training and inference processes}. The generator is trained in the latent space to reconstruct images at any resolution and aspect ratio. During inference, it can be applied to improve various diffusion models, enabling them to generate images of arbitrary resolution across various tasks.}
    \label{fig:framework}
\end{figure*}

\subsection{Preliminary Background}
\label{Sec:preliminary}
\paragraph{Variational Auto-Encoders.} 
VAEs~\cite{kingma2013auto} are advanced generative models that combine deep learning with probabilistic frameworks to learn efficient latent representations of data. These models are commonly employed as image tokenizers, offering a structured latent space that serves as the foundation for latent diffusion models~\cite{rombach2022high, ma2024sit, podell2023sdxl}.

Let $x \in \mathbb{R}^{3 \times H \times W}$ represent an RGB image, where $H$ and $W$ are height and width. A typical VAE consists of two main components: the encoder and decoder, which work together seamlessly to process and reconstruct data. The encoder transforms data $x$ into a latent space $z$ that obeys a Gaussian distribution. This mapping is crucial as it compresses the input into a lower-dimensional space,
\begin{equation}
q_\phi(z|x) = \mathcal{N}(z; \mu_\phi(x), \sigma^2_\phi(x)I),
\end{equation}
where \(\mu_\phi(x)\) and \(\sigma^2_\phi(x)\) are neural networks parameterized by \(\phi\). This ensures essential features are captured, facilitating effective reconstruction.

Once the data is encoded, the decoder takes over to reconstruct the input data from the latent space. It models the distribution of the data given the latent variables, enabling the generation of new samples,
\begin{equation}
p_\theta(x|z) = \mathcal{N}(x; \mu_\theta(z), \sigma^2I),
\end{equation}
where \(\mu_\theta(z)\) is a neural network parameterized by \(\theta\).

\paragraph{Implicit Neural Representation.}
INRs~\cite{sitzmann2020implicit} are usually adopted in the Coordinate-Based Networks modeling continuous signals. For images, the network fits a continuous function $f$ from pixel coordinates \((u, v)\) to RGB color,
\begin{equation}
 f_\theta(u, v) = \text{NN}((u, v); \theta),
 \label{eq:pre_INR}
\end{equation}
where \( \theta \) represents the network parameters. This formulation allows the network to map discrete pixels to continuous space, capturing intricate details in a generalized way.

\subsection{InfGen: Fixed Latent in, Arbitrary Image Out}
\label{sec: Arbitrary Image Out}
InfGen is a secondary generation model, in which the diffusion model is fundamentally viewed as a content generator, primarily generating fixed-compact content latent $z$. Furthermore, InfGen extends $z$ into images of arbitrary sizes, focusing on refining details and textures during this process.
There are two notable benefits for this paradigm: 1) \emph{high inference speed}: InfGen achieves very fast inference speeds by avoiding multi-step denoising on high-resolution latent spaces. This efficiency is due to the generating of arbitrary-resolution images on a compact latent $z$ with one-step inference, which reduces computational demands.
2) \emph{Plug and play}: InfGen offers strong generalization as it can be applied to any diffusion model trained on the same latent space. This flexibility makes it adaptable to various generative models and tasks without extra training.

\subsubsection{Overall Introduction of InfGen}
To enable InfGen to integrate seamlessly with various diffusion-based generative models and enhance their resolution flexibility, we propose a pipeline grounded in the latent space of the VAE. As depicted in Fig.~\ref{fig:framework}, this approach leverages the vanilla VAE architecture introduced in Section~\ref{Sec:preliminary}. Notably, our InfGen framework refrains from altering the VAE encoder, thereby preserving compatibility with existing diffusion models (e.g., SDXL~\cite{podell2023sdxl}, DiT~\cite{peebles2023scalable}). Instead, we incorporate a secondary generator within the decoder.

\vspace{-12pt}
\paragraph{Training pipeline.}During the training phase, high-resolution images are initially subjected to cropping and resizing operations to conform to a fixed dimension, such as $512\times512$. These preprocessed images are subsequently encoded by the VAE encoder into a fixed-size latent $z$, typically reduced by $8\times$, resulting in dimensions like $4\times64\times64$. The InfGen model is designed to achieve arbitrary resolution generation. The mapping is defined as:
\begin{equation}
    f: \boldsymbol{InfGen}(z, (h,w)) \rightarrow x_{(h, w)},
\end{equation}
where \(h, w\) is a dynamic parameter representing the expected image size, and \(x_{(h, w)}\) is the desired output image. 

\paragraph{Optimization objectives.}
The loss functions include adversarial loss, reconstruction loss, and perceptual loss, which is defined as: 
\begin{equation}
L_{AE} = \ell_1(x, \hat{x}) + \lambda_P \mathcal{L}_P(x, \hat{x}) + \lambda_G \mathcal{L}_G(\hat{x}),
\label{eq:optimization}
\end{equation}
where \(\ell_1\) represents $L_1$ loss between the reconstruction and the ground truth, \(\mathcal{L}_P\) is the perceptual loss measured by LPIPS~\cite{zhang2018unreasonable}, and \(\mathcal{L}_G\) is the adversarial loss derived from a PatchGAN discriminator~\cite{isola2017image}. 

\paragraph{Inference pipeline.} As illustrated in Fig.~\ref{fig:framework}, during the inference phase, by replacing the input latent vector with one generated by compatible diffusion models including but not limited to DiT~\cite{peebles2023scalable}, SDXL~\cite{podell2023sdxl}, SiT~\cite{ma2024sit}, FiT~\cite{lu2024fit}, InfGen can generate images of any size for a latent representation. 
Through this paradigm, InfGen can cost-effectively enhance the flexibility and resolution of diffusion models on different tasks, like class-guidance generation~\cite{Zheng2024MaskDiT,ma2024sit}, text-conditional generation~\cite{rombach2022high,podell2023sdxl}, and inpainting~\cite{zhang2023adding,corneanu2024latentpaint}.

\subsubsection{Arbitrary-Resolution Decoder Architecture}
As illustrated in~\cref{fig:framework} (a), we propose a novel architecture that enables reconstruction at any resolution. Building upon the traditional VAE structure, we introduce a transformer-based latent generator. Specifically, we treat the latent variables as prompts, utilizing them as keys and values.

Guided by the latent \( z \), we create a mask token serving as the query, which is tailored to the desired image dimensions \((h, w)\). For generating an image of size \((h, w)\), the mask token is shaped as \((\lceil h/8 \rceil, \lceil w/8 \rceil)\). This forms a key-query-value triplet suitable for the Multi-Head Self-Attention (MHSA) mechanism in transformer blocks.


Within the multi-layer transformer blocks, this triplet undergoes cross-attention computation. The mask token, acting as the query, continuously interacts with the latent keys to gather information through similarity calculations. Finally, we send the mask token to the decoder for upsampling, resulting in the arbitrary final image.

\subsubsection{Implicit Neural Positional Embedding}
Positional encoding is crucial for matching the spatial information between the mask and latent tokens. 
The image is split into fixed-size patches~\cite{khan2022transformers}, treating the position embedding for each patch as a learnable token~\cite{dosovitskiy2020vit}. Positional encoding helps retain spatial information. However, this typically restricts input image size, as these encodings are designed for a fixed number of tokens.

To overcome this limitation, we propose an Implicit Neural Positional Embedding (INPE) method to generate when the number of mask tokens is dynamic, which allows interactions between fixed-size latent representations and differently-sized mask tokens in cross-attention, enabling the regeneration of latent tokens with varying dimensions. 

\paragraph{Standardization and Conversion.}First, the coordinates of each mask token \((x^m, y^m)\) and latent token \((x^l, y^l)\) are standardized to map the different sizes to a unified scale:

\begin{equation}
(\hat{x}^m, \hat{y}^m) = (\frac{x^m}{W^m}, \frac{y^m}{H^m}),
(\hat{x}^l, \hat{y}^l) = (\frac{x^l}{W^l}, \frac{y^l}{H^l}), 
\end{equation}
where \(W_i\) and \(H_i\) denote the width and height of the latent token or the mask token. The standardized 2D coordinates are then converted to 3D Cartesian coordinates on the unit sphere: \(x = \cos(\pi \hat{y}_i) \cos(2\pi \hat{x}_i),  y = \cos(\pi \hat{y}_i) \sin(2\pi \hat{x}_i),  z = \sin(\pi \hat{y}_i)\). This mapping leverages spherical geometry to capture complex spatial relationships, enhancing continuity for smooth feature learning, reducing coordinate bias with symmetric structure. 

\paragraph{Fourier Transformation and Neural Mapping.}The 3D coordinates are transformed into high-frequency Fourier features to enhance the model's pattern-capturing ability:
\begin{equation}
\gamma(x, y, z) = [\cos(B[x, y, z]^T), \sin(B[x, y, z]^T)]
\end{equation}
where a diagonal matrix \(B\) is randomly sampled from a Gaussian distribution $\mathcal{N}\left ( \mu ,\sigma  \right ) $. It is used to map coordinates into higher-dimensional space. These Fourier features are then fed into an implicit neural network to generate dynamic positional encodings through Equation \ref{eq:pre_INR}.


\paragraph{Alignment and optimization.}

The generated positional encodings align with latent tokens and mask tokens to enhance information interaction in cross-attention. Dynamic encodings are integrated into \(Q\) and \(K\) to improve alignment and attention effectiveness. Finally, parameters \(\theta\) are synergistically optimized by minimizing the loss function introduced in~\cref{eq:optimization}.

\subsection{Training-free Resolution Extrapolation}
\label{sec: training}

To produce images beyond the training resolution, we introduce an iterative generation method, which is a training-free extrapolation method to scale the generated latent into arbitrary ultra-high resolution (e.g. 4K) continuously.

\begin{table}[h]
\small
    \centering
    \resizebox{7cm}{!}{
    \begin{tabular}{ccccc}
        \toprule
        Latent size & \makecell{Training \\ resolution}& \makecell{Reliable \\ extrapolation}&  \makecell{Max.\\$(s^h_i \times s^w_i)$}  \\
        \midrule
        $32\times32$ & $256 \sim 512$ &  $256 \sim 1024$  & $16 \times$ \\
        \midrule
        \multirow{2}{*}{$64\times64$} 
         & $512 \sim 1024$ & $512 \sim 2048$ &  $16 \times $ \\
         & $512 \sim 2048$ & $512 \sim 4096$ &  $64 \times$ \\
        \bottomrule
    \end{tabular}
    }
    \caption{\textbf{Recommended extrapolation resolution.}}
    \label{tab:extrapolation}
\end{table}

\vspace{-10pt}
\paragraph{Guaranteed base resolution.}
As shown in~\cref{tab:extrapolation}, InfGen receives a very-low resolution latent representation \( L \) of size \( 64 \times 64 \) and is capable of generating from the latent space to an image resolution \( R \), where \( R = s^h 512 \times s^w 512 \).
 An initial latent \( L_0 \) of size \( 64 \times 64 \) is used to generate an image \( I_1 \) with a resolution up to \( 2048 \times 2048 \):
\[
I_1 = \text{InfGen}(L_0, k_0^s) = \text{InfGen}(L_0, (s^h_0, s^w_0)),
\]
where \( s_h\) and \(s_w\) are scaling factors for height and width within the range \( 1 \leq s^h, s^w \leq 2 \).

\paragraph{Iterative extrapolated resolution.} The generated image is then encoded back into a latent representation for a further generation:
\begin{equation}
L_{n} = \text{Encoder}(I_{n-1}), 
I_{n} = \text{InfGen}(L_{n}, k^{s}_{n}),
\end{equation}
where each iteration involves a scaling factor \( k^s_{n} \). It is recommended to iterate within the scale range shown in~\cref{tab:extrapolation}.

\paragraph{Achieving arbitrary resolutions.} By repeating this process, the model achieves a final resolution \( R_f \),
\[
R_f =512\cdot \prod_{i=1}^{n} s^h_i \times 512 \cdot \prod_{i=1}^{n} s^w_i.
\]
The results in~\cref{tab:performance} verify this is a robust extrapolation method for flexible resolution enhancement, ensuring high-quality image generation.

\section{Experiment}
\label{sec:exp}
\paragraph{Dataset.} High-resolution training dataset is required to enhance the decoder's ability to express image details and textures during reconstruction. We selected 10 million images with resolutions exceeding \(1024^2\) from LAION-Aesthetic~\cite{schuhmann2022laion} as our training set. Further filtering resulted in a subset of 5 million images with resolutions over \(2048^2\), dividing the high-resolution training data into two parts. Since the input image resolution varies dynamically during training, we divide the training into two stages and change the batch size to avoid out-of-memory errors.

\paragraph{Implementation details.}During training, we keep the encoder of the pre-trained VAE frozen. For different batches, images are randomly cropped into different sizes. For inputs, they are then resized to a fixed size, such as $512\times512$, to obtain a fixed-size latent. For targets, they keep the cropped shape, except for performing another cropping to support decoding with arbitrary ratios. Due to the expensive computational and memory demands of high-resolution image reconstruction, the first training stage uses resolutions from \(512^2\) to \(1024^2\) with a batch size of 32. In the second stage, the resolution ranges from \(512^2\) to \(2048^2\), with the batch size reduced to 8. The training iterations are 500k and 100k respectively, lasting 15 days with 8 A100 GPUs. The AdamW optimizer~\cite{loshchilov2017decoupled} is used with an initial learning rate of $2e-4$, gradually decreasing to $1e-5$ using cosine decay. The \(\lambda_P\) and  \(\lambda_G\) in~\cref{eq:optimization} are both set to $0.1$.

\paragraph{Metrics.}For generation, we mainly focus on FID~\cite{heusel2017gans}, sFID~\cite{nash2021generating}, precision, and recall~\cite{kynkaanniemi2019improved}. Since FID requires downsampling to a resolution of 229$\times$229 for testing, downsampling can degrade the details of high-resolution images, making it unsuitable for evaluating the performance of generating high-resolution images. Therefore, we adopt the approach proposed in UltraPixel~\cite{ren2024ultrapixel} to crop high-resolution images into different 229$\times$229 patches for testing, denoted as FID$_{p}$, sFID$_{p}$, Pre.$_{p}$ and Rec.$_{p}$.
We also follow the previous tokenizer~\cite{podell2023sdxl} to report PSNR and SSIM as the metrics of reconstruction quality.

\begin{table}[h]
    \small
    \centering
    \begin{tabular}{cccccccc}
        \toprule
        Method & \multirow{2}{*}{\makecell{Input\&Output\\Resolution}}  & \multicolumn{3}{c}{ImageNet-50k} \\
        \cmidrule(lr){3-5}
        && rFID $\downarrow$ & PSNR$\uparrow$ & SSIM$\uparrow$\\
        \midrule
        $\text{VQGAN}^*$ & $256^2 \xrightarrow{}{256^2} $ & 4.99 & 20.00 & 0.629  \\
        VQGAN &  $256^2 \xrightarrow{}{256^2} $   & 1.19 & 23.38 & 0.762  \\
  
        SD-VAE &$256^2 \xrightarrow{}{256^2}$  & 0.74 & 25.68 & 0.820  \\
        
        SDXL-VAE&$256^2 \xrightarrow{}{256^2}$ & 0.68 & 26.04 & 0.834 \\
        \textbf{InfGen(Ours)}  &$256^2 \xrightarrow {}{256^2}$ & 1.07 &24.61&0.798 \\
        \textbf{InfGen(Ours)}  &$512^2  \xrightarrow {}{512^2} $ & 0.61 & 27.92 & 0.867\\
              \midrule
        
        SD-VAE &$256^2 \xrightarrow{}{512^2}$  & 1.43 & 24.14 & 0.759   \\
     
      
        \textbf{InfGen(Ours)}  &$256^2 \xrightarrow {}{512^2} $ &1.15 & 22.86& 0.728 \\

        
        \midrule
        & &\multicolumn{3}{c}{LAION-50k}\\
        \midrule
        
        \textbf{InfGen(Ours)}  &$256^2 \xrightarrow {}{256^2}$ & 3.30  & 26.14 &0.870\\
        \textbf{InfGen(Ours)}  &$512^2  \xrightarrow {}{512^2} $ &0.58  &28.85 &0.910 \\

        SD-VAE &$256^2 \xrightarrow{}{512^2}$  & 3.49  & 25.17 & 0.836  \\
        \textbf{InfGen(Ours)}  &$256^2 \xrightarrow {}{512^2} $&2.27&23.40&0.809 \\

        \textbf{InfGen(Ours)}  &$512^2 \xrightarrow {}{1024^2} $  & 1.16 &27.42& 0.872  \\
        \bottomrule
    \end{tabular}
    
    \caption{\textbf{Comparisons with other image tokenizers.} The evaluations are conducted on $256\times256$ ImageNet 50k validation set and LAION-5B 50k set. All models are trained on part of the LAION-5B dataset except "*" is trained on ImageNet.}
    \label{tab:tokenizer}
\end{table}

\subsection{Comparison with Alternative Image Tokenizers}
We evaluate our tokenizer against the discrete image tokenizers, like VQGAN~\cite{esser2021taming}, and continuous image tokenizers, such as SD VAE~\cite{rombach2022high} and SDXL VAE~\cite{podell2023sdxl}. As detailed in~\cref{tab:tokenizer}, our tokenizer achieves competitive reconstruction performance compared with the commonly employed VAEs, even though InfGen is trained on a more complex task. Additionally, we tested it on the LAION dataset on different input\&output resolutions to assess image reconstruction quality. LAION images present more complex scenes, and our results align with those from the ImageNet validation set, demonstrating that our tokenizer effectively handles both object-centric and scene-centric images.


\subsection{Improving Performance for Diffusion Models}
\begin{table*}[ht]
\footnotesize
\centering
\begin{tabular}{l@{\hspace{15pt}}|@{\hspace{15pt}}l@{\hspace{15pt}}c@{\hspace{15pt}}c@{\hspace{15pt}}c@{\hspace{15pt}}|@{\hspace{15pt}}l@{\hspace{15pt}}c@{\hspace{15pt}}c@{\hspace{15pt}}r}
\toprule
\textbf{Method} &\textbf{FID$_p$} $\downarrow$ & \textbf{sFID$_p$} $\downarrow$ & \textbf{Pre.$_p$} $\uparrow$ & \textbf{Rec.$_p$} $\uparrow$  & \textbf{FID$_p$} $\downarrow$ & \textbf{sFID$_p$} $\downarrow$ & \textbf{Pre.$_p$} $\uparrow$ & \textbf{Rec.$_p$} $\uparrow$  \\
\midrule
latent: $32\times32$ &\multicolumn{4}{c}{ image: 512 $\times$ 512} & \multicolumn{4}{|c}{ image: 1024 $\times$ 1024} \\
\midrule
  DiT-XL/2~\cite{peebles2023scalable}          & 44.17&25.99&0.44&0.26   &61.52 &48.11&0.27&0.22 \\
  InfGen+DiT-XL/2       & \textbf{39.81} \lred{9.9\%} &\textbf{24.25}&\textbf{0.46}&\textbf{0.30}   &\textbf{41.75} \lred{32\%}&\textbf{26.64} &\textbf{0.51}&\textbf{0.40}\\
\cmidrule(lr){0-8}
 SiT-XL/2~\cite{ma2024sit}           & 40.89  & 30.38 &0.35 &0.30  &67.98&54.23&0.23&0.22 \\
 InfGen+SiT-XL/2        & {38.83}\lred{5.0\%}  & \textbf{27.59} &\textbf{0.38} &\textbf{0.33} &\textbf{46.50}\lred{36\%}&\textbf{30.37}&\textbf{0.48}&\textbf{0.39}  \\
\cmidrule(lr){0-8}

MDTv2~\cite{gao2023mdtv2}   & 42.36&28.11 & 0.38 & 0.29   &60.63&51.62&0.25&0.23   \\
InfGen+MDTv2    & \textbf{38.46}\lred{9.1\%}&\textbf{25.43}&\textbf{0.42}&\textbf{0.35}   &\textbf{40.53}\lred{33\%}&\textbf{28.98}&\textbf{0.48}&\textbf{0.41}  \\

\Xhline{3\arrayrulewidth} 
 latent: $64\times64$ &\multicolumn{4}{c}{  image: 1024 $\times$ 1024} & \multicolumn{4}{|c}{ image: 2048 $\times$ 2048}\\
 \midrule
  DiT-XL/2~\cite{peebles2023scalable}         & 43.17 &32.27&0.50&0.40   &64.87&59.18&0.45&0.33 \\
  InfGen+DiT-XL/2       &\textbf{39.06} \lred{9.5\%}&\textbf{23.13}&\textbf{0.54} &\textbf{0.43}  &\textbf{56.21} \lred{13.4\%}&\textbf{40.24}&\textbf{0.56}&\textbf{0.42}\\

\midrule
FiTv2-XL/2i~\cite{lu2024fit}         & 42.04&31.12&0.49&0.41  &66.95&59.82&0.45&0.29 \\
  InfGen+
FiTv2-XL/2       &\textbf{38.77} \textcolor{red!50!white}{\lred{7.8\%}}&\textbf{22.17} &\textbf{0.51}&\textbf{0.43}  &\textbf{61.56} \lred{8.1\%}&\textbf{38.61}&\textbf{0.56}&\textbf{0.39}\\

\midrule
SD1.5~\cite{rombach2022high}        & 21.58 &33.91&0.44&0.45  &55.30&74.58&0.36&0.38 \\
  InfGen+
SD1.5      &\textbf{16.92} \lred{21\%}&\textbf{24.52}&\textbf{0.49}&\textbf{0.49}  &\textbf{41.12} \lred{26\%}&\textbf{45.18}&\textbf{0.51}&\textbf{0.49}\\
\Xhline{3\arrayrulewidth} 
 latent: $64\times64$ &\multicolumn{4}{c}{extrapolation$\rightarrow$ image: 3072 $\times$ 3072 } & \multicolumn{4}{c}{ image: 3072 $\times$ 3072 }\\
 \midrule
  DiT-XL/2~\cite{peebles2023scalable}         & 77.84&47.57&0.44&0.20   &\multicolumn{4}{c}{SD1.5~\cite{rombach2022high} } \\
  InfGen+DiT       &\textbf{45.94} \lred{41\%}&\textbf{32.52}&\textbf{0.65}&\textbf{0.34}  &\multicolumn{4}{c}{InfGen+
SD1.5 } \\

\midrule
FiTv2-XL/2~\cite{lu2024fit}         & 79.30&48.47&0.43&0.19  &73.13&62.83&0.34&0.24 \\
  InfGen+
FiTv2-XL/2       &\textbf{45.72} \lred{42\%}&\textbf{29.45}&\textbf{0.62}&\textbf{0.36}  &\textbf{40.75} \lred{44.3\%}&\textbf{38.97}&\textbf{0.57}&\textbf{0.44}\\
\bottomrule
\end{tabular}
\caption{\textbf{Improved performance at arbitrary resolution for diffusion models.} Our InfGen can improve the performance of existing latent-based diffusion models on all metrics across different resolutions, especially significant improvement in high-resolutions.}
\label{tab:performance}
\end{table*}

\begin{table*}[ht]
\centering
\footnotesize
\begin{tabular}{l c c c cc | c c c cr}
\toprule
 \textbf{Method} & \textbf{FID$_P$} $\downarrow$ &\textbf{sFID$_P$} $\downarrow$& \textbf{Pre.$_P$} $\uparrow$ & \textbf{Rec.$_P$} $\uparrow$ & \textbf{Latency (s)} $\downarrow$& \textbf{FID$_P$} $\downarrow$ &\textbf{sFID$_P$} $\downarrow$& \textbf{Pre.$_P$} $\uparrow$ & \textbf{Rec.$_P$} $\uparrow$ & \textbf{Latency (s)} $\downarrow$ \\
\textbf{Resolution} &\multicolumn{5}{c}{1024 $\times$ 1024} &\multicolumn{5}{c}{2048 $\times$ 2048} \\
\midrule

 ScaleCrafter~\cite{he2023scalecrafter} & 55.36&{90.34}&0.45&0.41 & {7} &144.61 &242.76
&0.45 &\underline{0.59}&97\\
 Inf-DiT~\cite{yang2024inf} &48.48&106.45&0.47&\underline{0.56} & 50 &142.05&255.07&{0.53}&{0.55} & 255\\
 UltraPixel~\cite{ren2024ultrapixel} &{48.37} &\underline{85.74}&\underline{0.58}&0.42 & 11 & \underline{127.26}&\underline{221.14}&\underline{0.61}&{0.55}&{20}  \\
InfGen+SD1.5 &\underline{44.85}&98.16&{0.50}&{0.52}&\textbf{2.9+0.4}&{139.14}&{225.27}&{0.53}&0.53 &\textbf{2.9+1.9} \\

InfGen+SDXL-B-1 &\textbf{35.14}&\textbf{79.25}&\textbf{0.67}&\textbf{0.66}&\underline{5.4+0.4}&
\textbf{96.41}&\textbf{189.68}&\textbf{0.59}&\textbf{0.62} &\underline{5.4+1.9} \\

\bottomrule
\end{tabular}
\caption{\textbf{Quantitative comparison with other methods.} Our InfGen+SDXL-B-1 yields the best performance on different resolutions while achieving an extremely low latency. All latencies are tested on an A100 GPU device. \underline{Underline} means the second-best.}
\label{tab:benchmark}
\end{table*}

InfGen model is expected to enhance resolution for generative models. After training, InfGen can be used as a plugin to replace the VAE decoder of existing latent space-based generative models, especially those sharing the same VAE encoder as ours. To evaluate, we tested several classic and newly published generative models, including DiT-XL/2~\cite{peebles2023scalable}, SiT-XL/2~\cite{ma2024sit}, MaskDiT~\cite{Zheng2024MaskDiT},  MDTv2~\cite{ma2024sit}, and FiTv2~\cite{lu2024fit}. By replacing these models' VAE decoders with InfGen, the generated latent can be decoded into outputs of any resolution. The results in~\cref{tab:performance} detail the performance improvements InfGen brings to each generative model for resolution upgrades. For the original models, which cannot generate images at arbitrary resolutions, we upsample their outputs to the evaluation size. We assessed two different latent space sizes: $32\times32$ and $64\times64$.

\paragraph{Quantitative comparison.} For the 32x32 latent space, we evaluated performance when generating images at $512\times512$ (4$\times$ upscaling) and 1024$\times$1024 (16$\times$ upscaling). For the 64x64 latent space, we assessed performance improvements at 1024$\times$1024, 2048$\times$2048, and 3072$\times$3072 resolutions.
The experimental results show that InfGen significantly enhances the generative performance of all models at high resolutions. For instance, the FID of $3072\times3072$, InfGen achieved up to a \textcolor{red}{41\%} improvement on DiT~\cite{peebles2023scalable} and a \textcolor{red}{44\%} improvement on SD1.5~\cite{rombach2022high}. Across all five evaluated resolutions, the average improvements were \textcolor{red}{8\%, 34\%, 13\%, 16\%}, and \textcolor{red}{42\%}, respectively. These results indicate that InfGen not only enables existing diffusion models to generate at any resolution but also significantly improves generation quality at different resolutions.

\vspace{-16pt}

\begin{figure*}[htbp]
    \centering
    \includegraphics[width=1.0\linewidth]{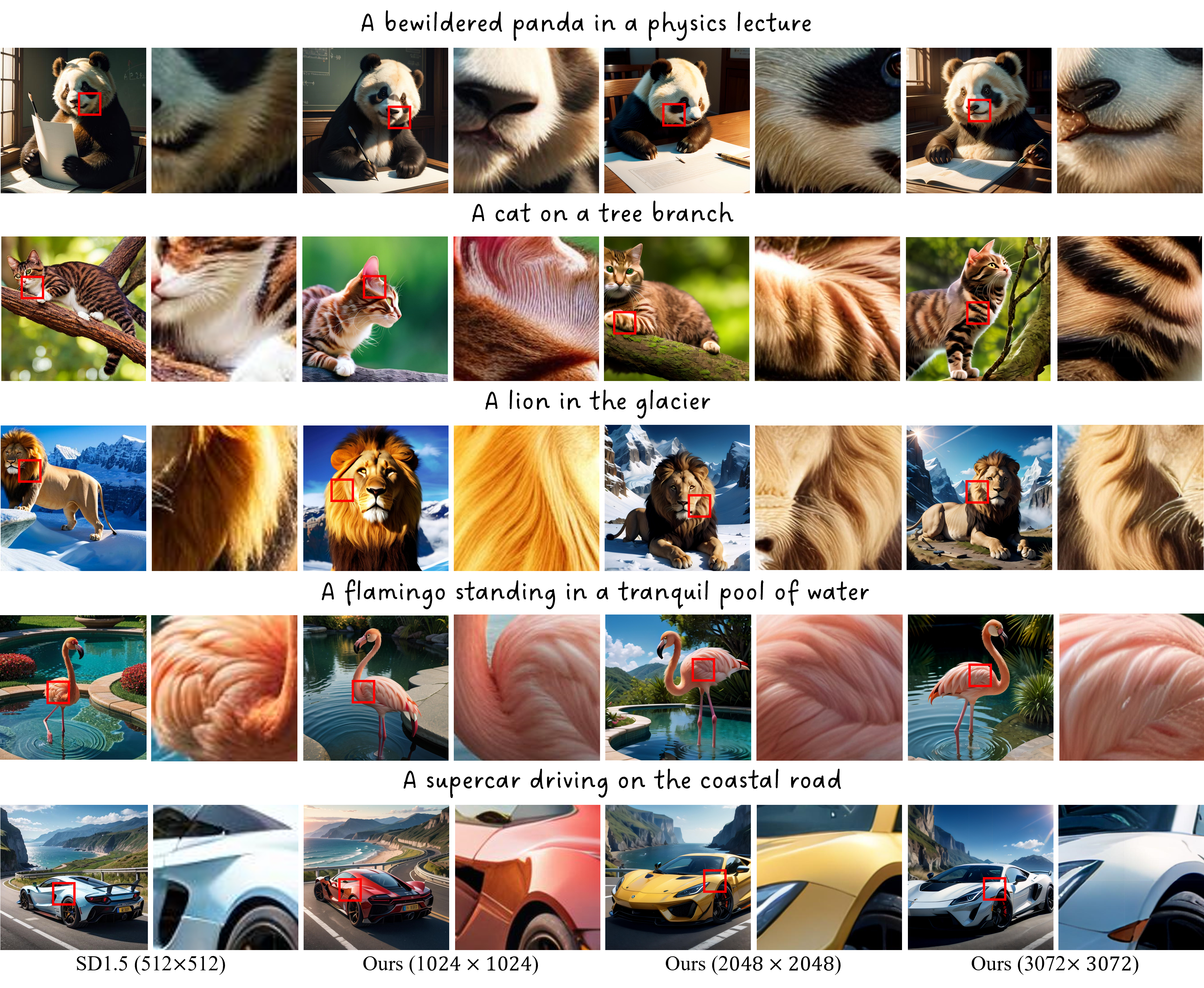}
    \caption{\textbf{Visualizations of arbitrary image generation.} The proposed InfGen improves the generation ability for LDMs~\cite{rombach2022high} across various resolutions. More visual examples are provided in the appendix.}
    \label{fig:vis}
    \vspace{-12pt}
\end{figure*}

\paragraph{Visualization comparison.}~\cref{fig:vis} presents a visual comparison between our InfGen model and the baseline across various resolutions. Using SD1.5 for high-resolution image synthesis results in visually unappealing structures and extensive irregular textures, which greatly reduce visual quality. In contrast, our method excels in producing superior semantic coherence and detailed intricacies. For example, at a resolution of 3072 × 3072, our generated images of a panda, cat, and lion display more detailed features. Even with a fixed-latent approach for generating images at different resolutions, our method consistently delivers visually pleasing and semantically coherent outcomes.

\subsection{Comparison to other State-of-the-Art Methods}
\paragraph{Benchmark and evaluation.}We also compare the text-to-image generation performance with some existing arbitrary resolution generation methods. Here we select recent methods, including training-free methods like ScaleCraft~\cite{he2023scalecrafter} and training-based methods, UltraPixel~\cite{ren2024ultrapixel} and Inf-DiT~\cite{yang2024inf}, for comparison. We comprehensively evaluate the performance of our model at resolutions of 1024$\times$1024 and 2048$\times$2048. For a fair comparison, we implement all methods on a same A100 GPU device and sample 1,000 images and 600 images for 1024$^2$ and 2048$^2$, respectively, where the reference images and captions are selected from the LAION dataset. Additionally, we include results for SDXL-Base-1.0~\cite{podell2023sdxl}, which achieves best performance with an FID$_P$ of 35.14 at 1024$\times$1024 resolution and 94.61 at 2048
$\times$2048 resolution, demonstrating its robustness in generating high-resolution images. These results further validate the effectiveness of our approach when integrated with diverse base models.

\vspace{-16pt}

\paragraph{Quantitative comparison.}As indicated in~\cref{tab:benchmark}, our method \texttt{InfGen}+SD1.5 performs competitively in terms of FID, sFID, Precision, and Recall in two high resolutions. Notably, \texttt{InfGen} has a significant disadvantage in inference efficiency, producing a 2048$\times$2048 image in $\sim$5 seconds, $4\times$ faster than the UltraPixel. Training-free method~\cite{he2023scalecrafter} and the super-resolution-based method~\cite{yang2024inf} take several minutes to generate a 2K image. These findings underscore the effectiveness of our method in generating ultra-high-resolution images with exceptional efficiency.

\section{Conclusion}
InfGen offers a highly efficient framework for generating images at arbitrary resolutions, addressing the limitations of existing methods that focus on generating arbitrary resolution latent spaces in diffusion models, often resulting in significant delays and computational overhead. By training a secondary generative model in a compact latent space, InfGen decodes low-resolution latent into images of any resolution without altering the structure or training of existing diffusion models. Our experiments demonstrate that InfGen, as an off-the-shelf enhancement, can improve diffusion models for arbitrary resolutions. Compared to other specialized methods, InfGen achieves superior quality and significantly reduces inference time, generating a 4K image only takes $7.4$ seconds. This advancement highlights InfGen's potential to significantly enhance fast ultra-high-resolution image generation capabilities.

\section*{Acknowledgements}
This work was supported partially by the Shanghai Artificial Intelligence Laboratory.  

We also acknowledge funding support from Hong Kong-based sources, including the Hong Kong RGC General Research Fund (152169/22E, 152228/23E, 162161/24E), the Research Impact Fund (R5060-19, R5011-23), the Collaborative Research Fund (C1042-23GF), the NSFC/RGC Collaborative Research Scheme (62461160332 $\&$ CRS$\_$HKUST602/24), the Areas of Excellence Scheme (AoE/E-601/22-R), and the InnoHK initiative (HKGAI).

{
    \small
    \bibliographystyle{ieeenat_fullname}
    \bibliography{main}
}

\end{document}